\documentclass[sigconf]{acmart}

\copyrightyear{2022}
\acmYear{2022}
\setcopyright{acmlicensed}\acmConference[KDD '22]{Proceedings of the 28th ACM SIGKDD Conference on Knowledge Discovery and Data Mining}{August 14--18, 2022}{Washington, DC, USA}
\acmBooktitle{Proceedings of the 28th ACM SIGKDD Conference on Knowledge Discovery and Data Mining (KDD '22), August 14--18, 2022, Washington, DC, USA}
\acmPrice{15.00}
\acmDOI{10.1145/3534678.3539241}
\acmISBN{978-1-4503-9385-0/22/08}

\usepackage{times}
\usepackage[utf8]{inputenc} 
\usepackage{hyperref}       
\usepackage{url}            
\usepackage{booktabs}       
\usepackage{amsfonts}       
\usepackage{nicefrac}       
\usepackage{lipsum}
\usepackage{microtype}      
\usepackage{multirow}
  
\usepackage{xcolor}         

\usepackage[title]{appendix}
\usepackage{csquotes}
\usepackage{bbm}
\usepackage{amsthm}
\usepackage{graphicx} 
\usepackage{tablefootnote}
\usepackage{bm}
\usepackage[linesnumbered,ruled]{algorithm2e}
\usepackage{algorithmic}

\usepackage{multirow}
\usepackage{adjustbox}
\usepackage{threeparttable}

\usepackage{wrapfig}

\usepackage{thm-restate}
\usepackage{dsfont}

\usepackage{array}
\newcolumntype{P}[1]{>{\centering\arraybackslash}p{#1}}

\usepackage{amsmath,amsfonts,bm}

\def\eqref#1{equation~\ref{#1}}

\def\floor#1{\lfloor #1 \rfloor}
\def\1{\bm{1}}

\def\vc{{\bm{c}}}

\def\vv{{\bm{v}}}

\def\vx{{\bm{x}}}

\def\mA{{\bm{A}}}

\def\mD{{\bm{D}}}

\def\mI{{\bm{I}}}

\def\mM{{\bm{M}}}

\DeclareMathAlphabet{\mathsfit}{\encodingdefault}{\sfdefault}{m}{sl}
\SetMathAlphabet{\mathsfit}{bold}{\encodingdefault}{\sfdefault}{bx}{n}

\def\sC{{\mathbb{C}}}
\def\sD{{\mathbb{D}}}

\def\sP{{\mathbb{P}}}

\DeclareMathOperator*{\argmax}{arg\,max}
\DeclareMathOperator*{\argmin}{arg\,min}

\usepackage{hyperref}
\usepackage{url}

\AtBeginDocument{%
  \providecommand\BibTeX{{%
    \normalfont B\kern-0.5em{\scshape i\kern-0.25em b}\kern-0.8em\TeX}}}

\begin{document}

\title{Availability Attacks Create Shortcuts}


\author{Da Yu}
\authornote{The work was done when Da Yu was an intern at Microsoft Research Asia.}
\affiliation{%
  \institution{Sun Yat-sen University}
  \city{Guangzhou}
  \country{China}}
\email{yuda3@mail2.sysu.edu.cn}

\author{Huishuai Zhang}
\authornote{Huishuai Zhang and Jian Yin are the corresponding authors.}
\affiliation{%
  \institution{Microsoft Research Asia}
  \city{Beijing}
  \country{China}}
\email{huzhang@microsoft.com}

\author{Wei Chen}
\affiliation{%
  \institution{Chinese Academy of Sciences}
  \city{Beijing}
  \country{China}}
\email{chenwei2022@ict.ac.cn}

\author{Jian Yin}
\authornotemark[2]
\affiliation{%
  \institution{Sun Yat-sen University}
  \city{Guangzhou}
  \country{China}}
\email{issjyin@mail.sysu.edu.cn}

\author{Tie-Yan Liu}
\affiliation{%
  \institution{Microsoft Research Asia}
  \city{Beijing}
  \country{China}}
\email{tyliu@microsoft.com}
\renewcommand{\shortauthors}{Yu and Zhang, et al.}

\begin{abstract}
Availability attacks\footnote{More precisely, we investigate clean-label availability attacks. Some availability attacks inject malicious training samples instead of perturbing existing ones \citep{biggio2012poisoning}.}, which poison the training data with imperceptible perturbations, can make the data \emph{not exploitable} by machine learning algorithms so as to prevent unauthorized use of data. In this work, we investigate why these perturbations work in principle. We are the first to unveil an  important population property of the perturbations of these  attacks: they are almost \textbf{linearly separable} when assigned with the target labels of the corresponding samples, which hence can work as \emph{shortcuts} for the learning objective. We further verify that linear separability is indeed the workhorse for availability attacks. We synthesize linearly-separable perturbations as attacks and show that they are as powerful as the deliberately crafted attacks. Moreover, such synthetic perturbations are much easier to generate.  For example,  previous attacks need dozens of hours to generate perturbations for ImageNet while our algorithm only needs several seconds. Our finding also suggests that the \emph{shortcut learning} is more widely present than previously believed as deep models would rely on shortcuts even if they are of an imperceptible scale and mixed together with the normal features. Our source code is published at \url{https://github.com/dayu11/Availability-Attacks-Create-Shortcuts}.
\end{abstract}





\maketitle

\begin{figure}[h]
\centering
  \includegraphics[width=1.0\linewidth]{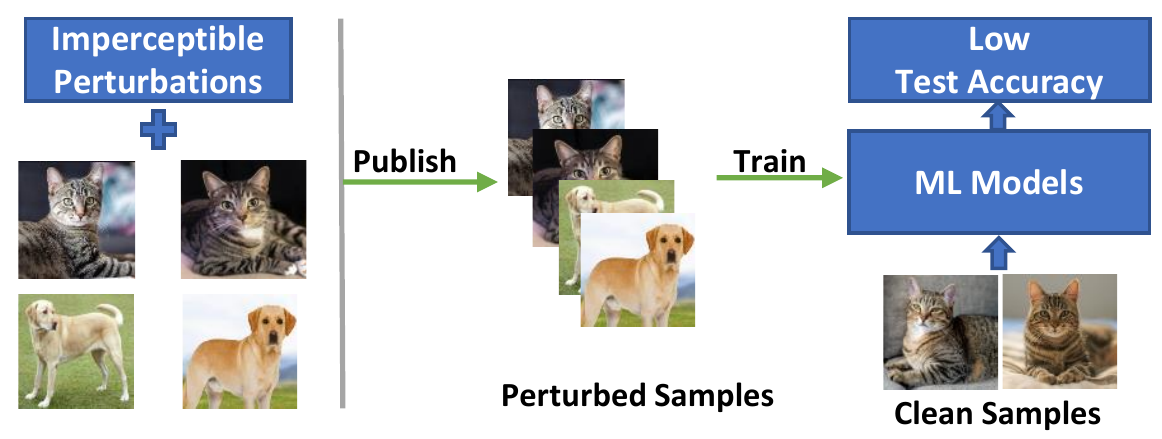}
  \caption{An illustration of clean-label availability attacks. }
  \label{fig:unlearnable_setting}
\end{figure}

\section{Introduction}



Sharing personal data online has become an important lifestyle for many people. Despite big datasets crawled from the Internet keep advancing the state-of-the-art deep models \citep{devlin2018bert, he2020momentum, chen2020big}, there are increasing concerns about the unauthorized use of personal data \citep{hill2019photos, prabhu2020large, carlini2020extracting}. For instance, a private company  has collected more than three billion face images to build commercial face recognition models without acquiring any user consent \citep{hill2020secretive}. To address those concerns, many data poisoning attacks have been proposed to prevent  data from being learned by unauthorized deep models \citep{feng2019learning,shen2019tensorclog,huang2021unlearnable,yuan2021neural,fowl2021preventing, fowl2021adversarial,tao2021provable}. They  add imperceptible perturbations to the training data so that the model cannot learn much information from the data and the model accuracy on unseen data is arbitrarily bad. These attacks make the data not \emph{available/exploitable} by machine learning models and are known as \emph{availability attack} \citep{biggio2018wild}. We give an illustration of this type of attack in Figure~\ref{fig:unlearnable_setting}. 




In literature, there are roughly three methods to construct the availability attack against deep neural networks. The first method generates the perturbations as the solution of a bi-level optimization problem \citep{biggio2012poisoning,feng2019learning,fowl2021preventing,yuan2021neural}. The bi-level optimization problem updates the perturbations to minimize the loss  on perturbed data while maximizing the loss on clean data. 

Secondly, \citet{huang2021unlearnable} conceive a simpler poisoning attack called \emph{error-minimizing noise}, where the perturbation on training data is crafted by minimizing the training loss. The intuition is that if the perturbation can reduce the loss to zero,  then there is nothing left for backpropagation in the regular training procedure. Recently,  \citet{nakkiran2019discussion} and \citet{fowl2021adversarial} point out that \emph{error-maximizing noises}, which are commonly used as adversarial examples, can serve as an availability attack as well.  Despite these quite different approaches,  all of them are powerful availability attacks. Intrigued by this observation, we ask the following question:

\begin{center}
\emph{What is the underlying workhorse for  availability attacks against deep neural networks?}
\end{center}

\begin{figure*}[t]
\centering
  \includegraphics[width=0.8\linewidth]{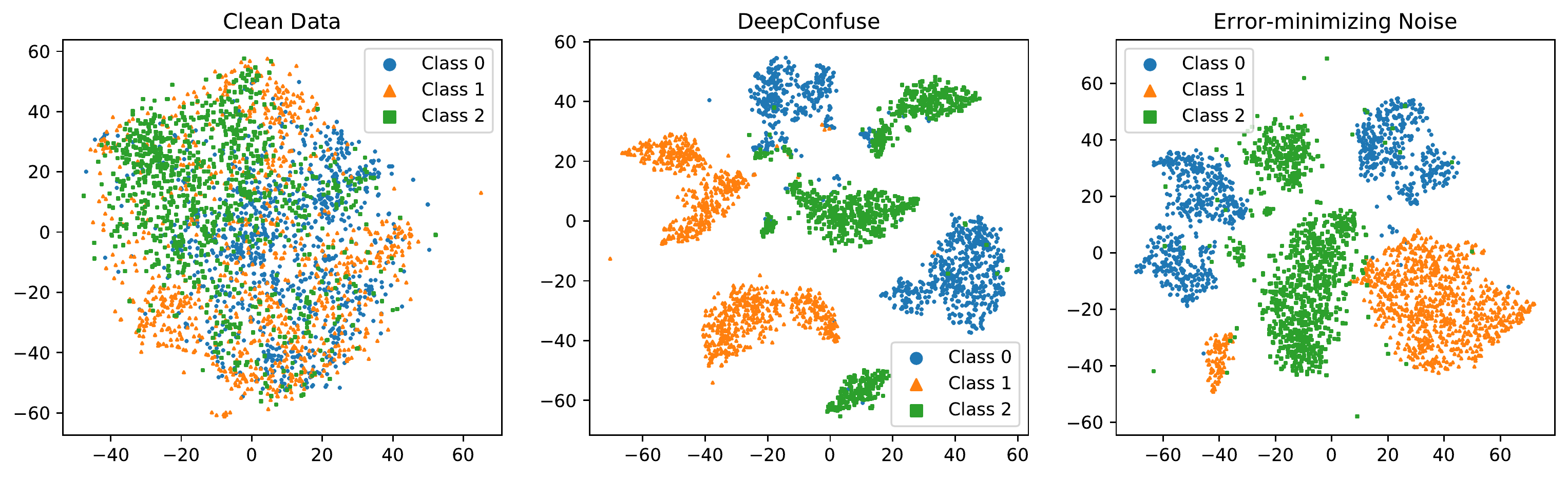}
  \caption{T-SNEs of the first three classes of clean CIFAR-10 data and the perturbations generated via DeepConfuse \citep{feng2019learning} and error-minimizing noises \citep{huang2021unlearnable}. The perturbations are flattened and normalized into unit norms. }
  \label{fig:tsne}
\end{figure*}

To answer this question, we first take a close look at the perturbations of existing attacks. We visualize the perturbations of several availability attacks via two-dimensional T-SNEs \citep{van2008visualizing} in Figure~\ref{fig:tsne} and Figure~\ref{fig:add_tsne} in Appendix~\ref{apdx:more_tsne}. The experimental setup is depicted in Section~\ref{sec:draw_tsne}.  Surprisingly, the perturbations with the same class label are well clustered, suggesting that the perturbations would be nearly \textbf{linearly separable} in the original high-dimensional space. We confirm this by fitting the perturbations with linear models. The perturbations are assigned with the labels of their target examples. It turns out that simple logistic regression models can fit the perturbations of four representative attacks with $>90\%$ training accuracy. 

Conceptually, the reason why current availability attacks work could be that the imperceptible perturbations create a kind of \emph{shortcut}. They are so simple, i.e., linearly-sperarable, that the deep models tend to rely on them to make prediction while ignoring the true features. This extends the concept of existing shortcut learning \citep{geirhos2020shortcut}, which is often referred to that deep models tend to rely on some natural features, correlated with labels but not causal ones, e.g., ``grass'' is a shortcut for recognizing ``cow'' in natural images \citep{beery2018recognition,geirhos2020shortcut}. In contrast, we expose a more explicit form of shortcuts and one may create such shortcut intentionally. 

To further confirm that creating shortcut is also sufficient (not only necessary) for a successful availability attack, we reverse the above procedure: synthesizing some simple linearly-separable perturbations to see if they can serve as availability attacks. Specifically, we first generate some initial synthetic perturbations via a method in  \citet{guyon2003design} and then add a new post-processing procedure  so that the synthetic perturbations remain effective when  data augmentations are applied. Extensive experiments on benchmark datasets and models demonstrate that synthetic perturbations can be as powerful as existing availability attacks. Notably, generating synthetic perturbations is significantly easier and cheaper than existing attacks as it does not require solving any optimization problems.  For example, recent attacks need dozens of hours to generate perturbations  for the ImageNet data, while generating synthetic perturbations only needs several seconds. This finding reveals that one can instantiate a successful availability attack by just creating shortcuts. 

This paper unveils that deep learning models would overwhelmingly rely on spurious shortcuts even though the shortcuts are scaled down to an imperceptible magnitude, which exposes a fundamental vulnerability of deep models.  Our contributions  are summarized as follows:
\begin{itemize}
    \item We reveal that the perturbations of several existing availability attacks are (almost) linearly separable. 
    \item We propose to use synthetic shortcuts to perform availability attack, which is much cheaper and easier to conduct.
    \item We link availability attacks with   shortcut learning and greatly widen the understanding of shortcuts in deep learning.  
\end{itemize}

\subsection{Related Work}

\textbf{Data poisoning.} In general, data poisoning attacks perturb training data to intentionally cause some malfunctions of the target model \citep{biggio2018wild,goldblum2020dataset,schwarzschild2021just}.  A common class of poisoning attacks aims to cause test-time error on some given samples \citep{shan2020fawkes,geiping2020witches,cherepanova2021lowkey,zhang2021data} or on all unseen samples \citep{feng2019learning,shen2019tensorclog,huang2021unlearnable,yuan2021neural,fowl2021preventing, fowl2021adversarial}. The latter attacks are also known as availability attacks \citep{barreno2010security}.  In this work, we investigate  and reveal the workhorse of availability attacks. We show that the perturbations of these availability attacks are (almost) linearly separable. We further confirm that synthesised  linearly-separable perturbations can perform strong attacks.

Backdoor attacks are another type of data poisoning attack that perturbs training data so that the attacker can  manipulate the target model's output with a designed trigger \citep{shafahi2018poison,nguyen2020input,saha2020hidden,tang2020embarrassingly,nguyen2021wanet,doan2021backdoor}. The perturbations of backdoor attacks have two major differences compared to those of availability attacks. Firstly, the perturbations of availability attacks are  imperceptible. Secondly, advanced availability attacks   use a different perturbation for every sample while a backdoor trigger is applied to multiple samples. In the threat model of availability attacks, the data are probably crowdsourced. It is preferred to use different perturbations for different samples in such a setting. Otherwise, the adversarial learner can remove the perturbation from all related samples if any of the poisoned images are leaked.

\textbf{Shortcut learning.} Recently, the community has realized that deep models may rely on shortcuts to make decisions \citep{beery2018recognition,niven2019probing,geirhos2020shortcut}. Shortcuts are spurious features that are correlated with target labels but do not generalize on test data.  \citet{beery2018recognition} show that a deep model would fail to recognize cows when the grass background is removed, suggesting that the model takes ``grass'' as a shortcut for ``cow''. \citet{niven2019probing} show that large language models use the strong correlation between some simple words and labels to make decisions, instead of trying to understand the sentence. For instance, the word ``not'' is directly used to predict negative labels.  In this work, we show  shortcut learning exists more widely  than previously believed. Our experiments in Section~\ref{sec:syntehtic_noise} demonstrate that deep models only pick shortcuts even if the shortcuts are scaled down to an imperceptible magnitude and mixed together with normal features.  These experiments reveal another form of shortcut learning, which has been unconsciously exploited by availability poisoning attacks. There also exist other synthesized datasets that offer a stratification of features \citep{hermann2020shapes,shah2020pitfalls}. Those synthetic data contain shortcuts that can not be used as perturbations as they are visible and affect the normal data utility. For example, \citet{shah2020pitfalls} generate synthetic data by vertically concatenating images from the MNIST and CIFAR-10 datasets.

 

\subsection{Notations}

We use bold lowercase letters, e.g., $\vv$, and bold capital letters, e.g., $\mM$, to denote vectors and matrices, respectively. The $L_{p}$ norm of a vector $\vv$ is denoted by $\|\vv\|_{p}$. A sample consists of a feature vector $\vx$ and label $y$. We use $\sD$ to denote a dataset that is sampled from some data distribution $\mathcal{D}$. In this paper, we focus on classification tasks. The classification loss of a  model $f$ on a given sample is denoted by $\ell(f(\vx), y)$.

\section{Availability Attacks Use Linearly-Separable Perturbations}
\label{sec:why}

In this section, we investigate the common characteristic of existing availability attacks. First, We briefly introduce three different approaches to construct availability attacks. Then, we  visualize the perturbations of advanced attacks with two-dimensional T-SNEs. The plots suggest that the perturbations of all three kinds of attacks are some `easy' features. Finally, we verify that the perturbations of these attacks are almost linearly separable by fitting them with simple models.

\subsection{Three Types Of Availability Attacks}
\label{sec:three_attacks}

\subsubsection{The Alternative Optimization Approach}

We first introduce the alternative optimization approach to generate perturbations for availability attacks. It solves the following bi-level objective,  

\begin{equation}
\begin{aligned}
\label{eq:bi_level}
&\argmax_{\{\boldsymbol\delta\}\in \Delta} \mathbb{E}_{(\vx,y)\sim \mathcal{D}}[\ell(f^{*}(\vx), y)],\\
&\text{s.t. } f^{*}\in \argmin_{f} \sum_{(\vx,y)\in\sD} \ell(f(\vx+\boldsymbol\delta), y),
\end{aligned}
\end{equation}
where $\boldsymbol\delta$ is a sample-wise perturbation and $\Delta$ is a constraint set for  perturbations.  The two formulas in Equation~\ref{eq:bi_level} directly reflect the goal of availability attacks. Specifically, the optimal solution on perturbed data (specified by the second formula) should have the largest loss on clean data (specified by the first formula). The constraint set $\Delta$ is  set to make the perturbations imperceptible, e.g., a small $L_{p}$ norm ball. 



Directly solving Equation~(\ref{eq:bi_level}) is intractable for deep neural networks.  Recent works have designed multiple approximate solutions \citep{feng2019learning,fowl2021preventing,yuan2021neural}. \citet{feng2019learning} use multiple rounds of optimization to generate perturbations. At each round, they first approximately optimize the second objective by  updating a surrogate target model on perturbed data for a few steps. Then they approximately optimize the first objective by updating a generator for a few steps. The outputs of the generator are used as perturbations. Another example is the Neural Tangent Generalization Attacks (NTGAs) in \citet{yuan2021neural}. They approximately optimize the bi-level objective based on the recent development of Neural Tangent Kernels \citep{jacot2018neural}.

\subsubsection{The Error-minimizing Noise}

\citet{huang2021unlearnable} propose another bi-level objective to generate perturbations. Instead of solving Equation~(\ref{eq:bi_level}), they use the following objective, 


\begin{equation}
\begin{aligned}
\label{eq:error_min}
\argmin_{\{\boldsymbol\delta\}\in\Delta} \mathbb{E}_{(\vx,y)\sim\mathbb{D}} [\min_{f}\ell(f(\vx+\boldsymbol\delta), y)].
\end{aligned}
\end{equation}

The perturbations are intentionally optimized to reduce the training loss. The main motivation is that if the training loss is zero, then the target model will have nothing to learn from the data because there is nothing to backpropagate. A randomly initialized model is used as a surrogate of the target model. They also use multiple rounds of optimization to generate perturbations. At each round, they first train the surrogate model for a few steps to minimize the loss on perturbed data. Then they optimize the perturbations to also minimize the loss of the surrogate model. They repeat the above process until the loss on perturbed data is smaller than a pre-defined threshold. 

\subsubsection{Adversarial Examples}

Instead of using bi-level objectives, \citet{fowl2021adversarial} show that the common objectives of adversarial examples are sufficient to generate powerful data poisoning perturbations. They use both  untargeted (the first objective) and targeted adversarial examples (the second objective),

\begin{equation}
\begin{aligned}
\label{eq:error_max}
&\argmax_{\{\boldsymbol\delta\}\in \Delta} \mathbb{E}_{(\vx,y)\sim\mathbb{D}} \left[\ell\left(f\left(\vx+\boldsymbol\delta\right), y\right)\right],\\
&\argmin_{\{\boldsymbol\delta\}\in \Delta} \mathbb{E}_{(\vx,y)\sim\mathbb{D}} \left[\ell\left(f\left(\vx+\boldsymbol\delta\right), y'\right)\right],
\end{aligned}
\end{equation}
where $y'\neq y$ is an incorrect label and $f$ is a trained model. Surprisingly, \citet{fowl2021adversarial} demonstrate that these simple objectives can generate perturbations that achieve state-of-the-art attack performance.

\subsection{Visualizing The Perturbations}
\label{sec:draw_tsne}

Although the three approaches in Section~\ref{sec:three_attacks} have different objectives, they all manage to perform powerful attacks. Intrigued by this observation, we try to find out whether there is a common pattern among different types of availability attacks. If so, the common pattern may be the underlying workhorse for availability attacks.

To find such a common pattern, we first visualize different types of perturbations, including DeepConfuse \citep{feng2019learning}, NTGA \citep{yuan2021neural}, error-minimizing noises \citep{huang2021unlearnable}, and adversarial examples \citep{fowl2021adversarial}. We compute their two-dimensional t-SNEs \citep{van2008visualizing}. These four attacks achieve advanced attack performance and cover all the three approaches in Section 2.1.  We use their official implementations to generate perturbations. Detailed configurations are in Appendix~\ref{apdx:exp_details}.


The two-dimensional t-SNEs of DeepConfuse and error-minimizing noises   are shown in Figure~\ref{fig:tsne}. The plots of NTGA and adversarial examples are presented in Figure~\ref{fig:add_tsne} of Appendix~\ref{apdx:more_tsne} due to the space limit.   Surprisingly, for all the attacks considered, the perturbations for the same class are well clustered, suggesting that even linear models can classify them well. For comparison purposes, we also compute the t-SNEs of the clean data. As shown in Figure~\ref{fig:tsne}, in contrast with the t-SNEs, the projections of different classes of the clean data are mixed together, which indicates that they require a complex neural network to be correctly classified. This observation suggests that using linearly-separable perturbations may be the common pattern among availability attacks.

\subsection{Availability Attacks Use Linearly-Separable Perturbations}

To quantify the `linear separability' of the perturbations, we fit the perturbations with simple models and report the training accuracy. The perturbations are labeled with the labels of the corresponding target examples. The simple models include linear models and two-layer neural networks. Details can be found in Appendix~\ref{apdx:exp_details}. We  choose  one-layer (linear) and two-layer neural networks because they are easy to implement in existing deep learning frameworks \citep{abadi2016tensorflow,paszke2019pytorch}. We note there are other choices that also fit the task, e.g.,  support-vector machines.




\begin{table}[t]
\caption{Training accuracy (in \%) of simple models on clean data and the perturbations of different attacks.}

\label{tbl:linear_acc}
\centering

\begin{tabular}{c|cc}

Algorithm & Linear Model & Two-layer NN  \\ \hline
Clean Data &    $49.9$            &  $70.1$ \\ 
 DeepConfuse \citep{feng2019learning} & $100.0$  & $100.0$ \\ 
  NTGA \citep{yuan2021neural} & $100.0$ & $100.0$ \\ 
 Error-minimizing \citep{huang2021unlearnable} & $100.0$ & $100.0$  \\ 
 Adv. Examples (Untargeted) \citep{fowl2021adversarial} & $91.5$ & $99.9$ \\ 
 Adv. Examples (Targeted) \citep{fowl2021adversarial} & $100.0$ & $100.0$ \\ 
\end{tabular}
\end{table} 

The results are presented in Table~\ref{tbl:linear_acc}.  Compared to the results on clean data,  simple models can easily fit the perturbations. On all attacks considered, linear models achieve more than $90\%$ training accuracy and two-layer neural networks achieve nearly $100\%$ training accuracy. These results confirm that the perturbations of advanced availability attacks are all (almost) linearly separable.

Existing attacks against deep neural networks all use ReLU activation functions in their crafting models. Deep models with ReLU activation functions are known to learn piecewise linear functions in input space \citep{arora2016understanding}. Therefore, it is natural to wonder whether the linear separability is stemming from the property of ReLU.  In Section~\ref{apdx:tanh}, we replace the ReLU layers with Tanh layers in the crafting models of adversarial examples and error-minimizing noises. It turns out that simple models still can easily fit the new perturbations, which demonstrates that the linear separability is an intrinsic property of these availability attacks rather than something associated with a specific network structure.



\subsection{Connecting To Shortcut Learning}


The fact that the perturbations can be easily fitted by linear models naturally connects to a recent concept named shortcut learning \citep{geirhos2020shortcut}. Shortcut learning summarizes a general phenomenon when any learning system makes decisions based on spurious features that do not generalize on realistic test data\footnote{\citet{geirhos2020shortcut} use a more specific definition of shortcuts. They denote shortcuts as those features that do not generalize on out-of-distribution (OOD) data. We note that poisoning attacks would change the distribution of training data and hence make the clean test data `OOD' with respect to the trained model.}. Shortcut features have been found in different fields. For vision tasks, \citet{beery2018recognition} show deep models fail to recognize cows when the grass background is removed from images, suggesting the grass background is a shortcut for predicting cows.  In the field of natural language processing, \citet{niven2019probing} show language models use the strong correlation between some simple words and labels to make decisions, instead of really understanding the data.





\begin{figure}[h]
\centering
  \includegraphics[width=0.65\linewidth]{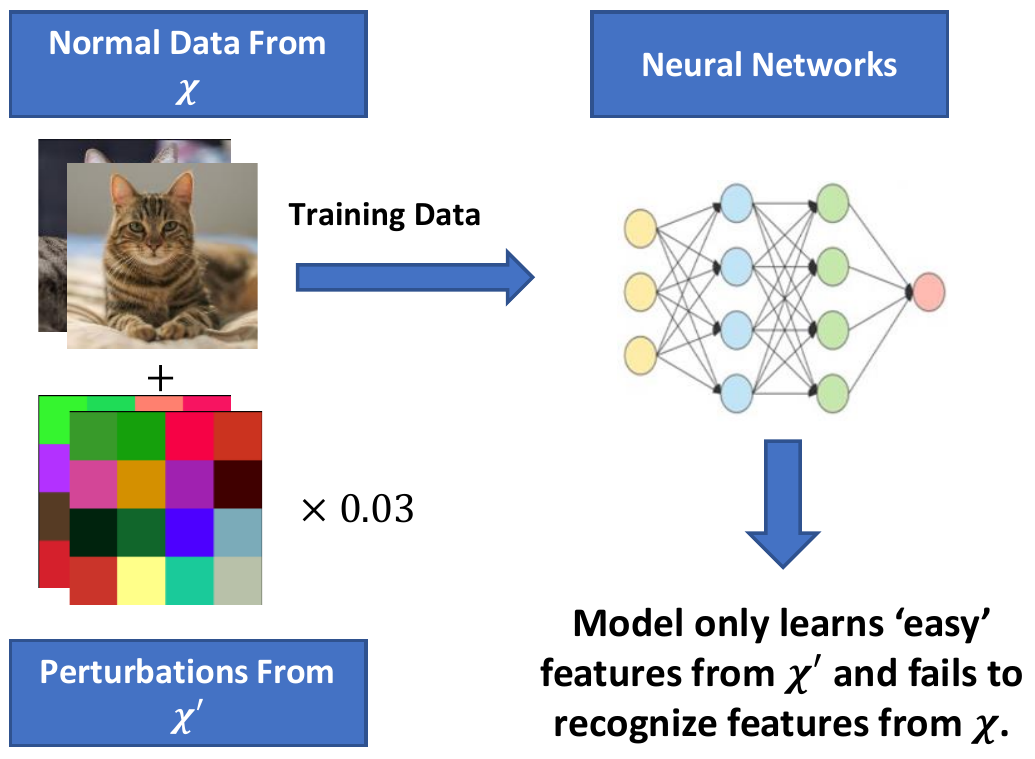}
  \caption{An illustration of  how the perturbations of availability attacks work as shortcuts.  }
  \label{fig:shortcut}
\end{figure}

With the presence of shortcut learning, it seems reasonable to postulate that the perturbations of existing attacks succeed by creating shortcuts to the target model. We give an illustration in Figure~\ref{fig:shortcut}. A major difference between the perturbations of poisoning attacks and existing shortcut features is that the perturbations are of an imperceptible scale and mixed together with useful features. Since there is no direct evidence to show deep models will take this kind of shortcuts, 
in the next section, we design experiments to confirm the postulated explanation.  We synthesize imperceptible and linearly-separable perturbations and show deep models are very vulnerable to such synthetic shortcuts.

\section{Linear Separability Is A Sufficient Condition For  Availability Attacks to Succeed}
\label{sec:syntehtic_noise}

Although we have demonstrated that the perturbations of four advanced attacks are all almost linearly separable, it is  a bit early to claim that `linear separability' is the underlying working principle of availability poisoning attacks. Perturbations are linearly separable  may only be a necessary but not sufficient condition for poisoning attacks to succeed. In order to verify this postulated explanation, we use simple synthetic data  to serve as perturbations and compare their effectiveness with existing poisoning attacks. It turns out that the synthetic perturbations  are as powerful as advanced attacks.  


The rest of this section is organized as follows. In Section~\ref{subsec:generation}, we first give an algorithm for generating synthetic data as perturbations. In Section~\ref{subsec:comparison}, we verify the effectiveness of synthetic perturbations on different models and datasets.

\subsection{Generating Synthetic Perturbations as Shortcuts}
\label{subsec:generation}

The synthetic perturbations in this section are generated via two building blocks. In the first  block, we use a method in \citet{guyon2003design} to generate samples from some normal distributions. In the second block, we transfer the  samples into the image format so that they can be applied  to benchmark vision tasks. We give the pseudocode in Algorithm~\ref{alg:generation}.

The first building block  proceeds as follows. We first generate some points that are normally distributed  around the vertices of a hypercube. The points around the same vertex are assigned with the same label. Then for each class, we introduce  different covariance. Any two classes of the generated points can be easily classified by a hyperplane as long as the side length of the hypercube is reasonably large.

In the second building block, we pad each dimension of the sampled points and reshape them into two-directional images. The padding operation introduces local correlation into the synthetic images.  Local correlation is an inherent property of natural images. In Section~\ref{apdx:no_pad}, we show the padding operation is necessary to make the synthetic  perturbations remain effective when data augmentation methods are applied.

\begin{algorithm}
\caption{Generating Perturbations for Vision Datasets}
   \label{alg:generation}
\begin{algorithmic} [1]
   \STATE {\bfseries Input:} number of classes $k$, number of examples in each class $\{n_{i}\}_{i=1}^{k}$, image size $(w,h)$,  patch size $p$, norm bound $\epsilon$.
   
   \medskip
   
   \STATE Compute $w'=\floor{w/p}+1$ and $h'=\floor{h/p}+1$. 
   
   \medskip
   //\textsl{ The first block: generate some initial data points.}
   
   \STATE Create arrays \{$\mD^{(i)}\in\mathbb{R}^{n_{i}\times w'h'}\}_{i=1}^{k}$ that will contain points in $w'h'$-dimensional hypercube.
   
   \FOR{$i=1$ {\bfseries to} $k$} 
			\STATE Initialize $\mD^{(i)}$ with samples from  $\mathcal{N}(0,\mI_{w'h'\times w'h'})$. 
            \STATE //\textsl{ Introduce random covariance among columns.}
			\STATE Uniformly sample the elements of $\mA\in\mathbb{R}^{w'h'\times w'h'}$ from $[-1,1]$.
			\STATE Compute $\mD^{(i)}=\mD^{(i)}\mA$.
			\STATE Randomly choose an unused vertex and let $\vc^{(i)}\in\mathbb{R}^{w'h'}$ be its coordinates.
			\STATE //\textsl{ Move the sampled points to the chosen vertex.}
			\STATE Compute $\mD^{(i)}=\mD^{(i)}+\vc^{(i)}$, i.e., $\vc^{(i)}$ is added to each row of $\mD^{(i)}$.
			\STATE Assign the rows of $\mD^{(i)}$ with label $i$. %
   \ENDFOR
   
   \medskip
   
  //\textsl{ The second block: duplicating each dimension to introduce local correlation.}
  \STATE Duplicate each dimension of the initial data points for $p^{2}$ times and reshape the results into two-dimensional $p\times p$ patches.
  \STATE Put the patches together and take crops to generate synthetic noises with size $(w,h)$.
  
  \medskip
   //\textsl{ Scale down the magnitude of synthetic data and harvesting perturbations.}
  \STATE Normalize each synthetic sample with $L_{2}$ norm bound $\epsilon$.
   \STATE Add perturbations to clean images with the same labels.

\end{algorithmic}
\end{algorithm}

\begin{figure*}
\centering
  \includegraphics[width=0.7\linewidth]{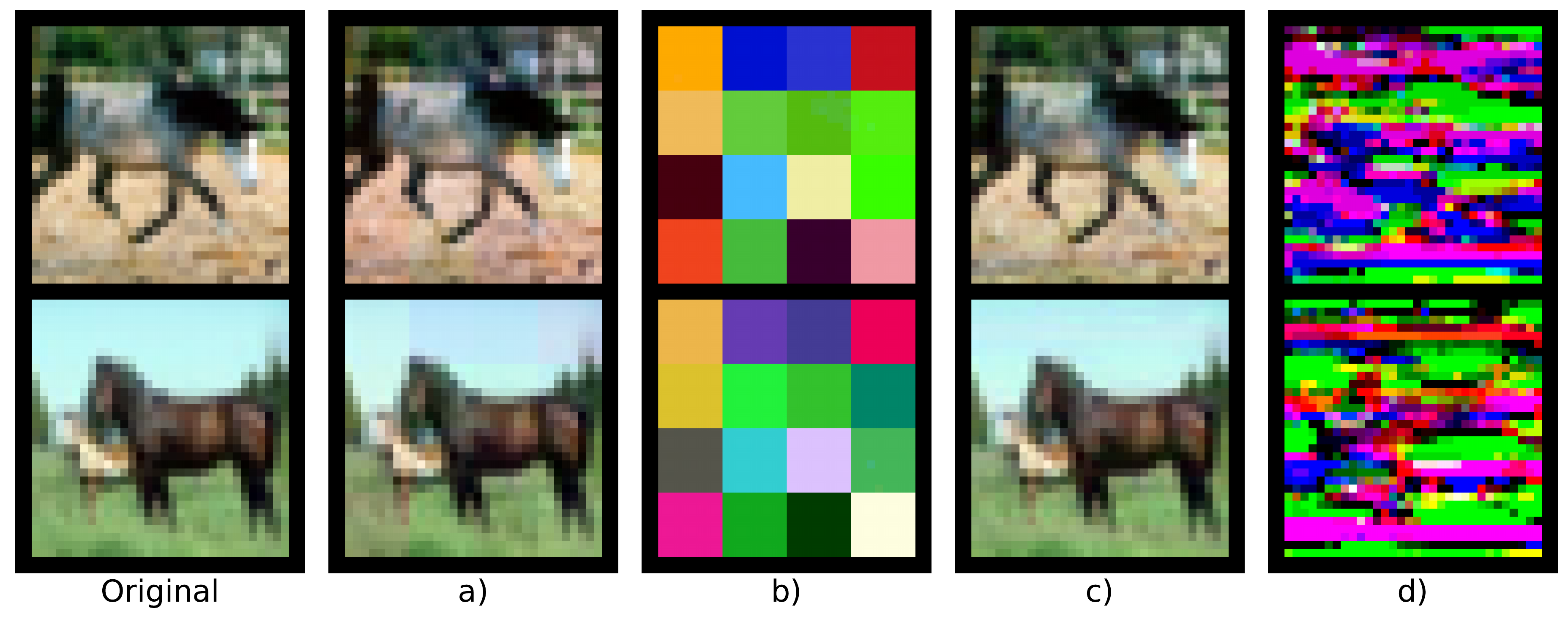}
  \caption{Visualization of perturbed images and normalized perturbations. Columns a) and b)  use synthetic perturbations. Columns c) and d) use the attack in  \citet{huang2021unlearnable}.  }
  \label{fig:synthetic_images}
\end{figure*}

 In Algorithm~\ref{alg:generation}, the synthetic images are scaled down before being used as perturbations.  We visualize the synthetic perturbations and corresponding perturbed images in Figure~\ref{fig:synthetic_images}. We also visualize the perturbations in \citet{huang2021unlearnable} for a comparison. The details of perturbations can be found in Section~\ref{subsec:comparison}. As shown in Figure~\ref{fig:synthetic_images}, the synthetic perturbations do not affect data utility.

\subsection{Synthetic Perturbations Are Highly Effective as Availability Attacks}
\label{subsec:comparison}



Now we verify the effectiveness of synthetic perturbations and make comparisons with existing availability attacks. We perturb the entire training set following the main setup in previous works \citep{feng2019learning,huang2021unlearnable,yuan2021neural,fowl2021adversarial}. That is, we synthesize a perturbation  for every training example. In Section~\ref{subsec:singleclass} and~\ref{apdx:different_percentage}, we show synthetic perturbations are still effective when only partial training data are poisoned. 




We use $L_{2}$-norm for synthetic perturbations to keep the sample-wise variation in the same class. We normalize the synthetic noises into a $L_{2}$-norm ball with radius $\sqrt{d}\epsilon'$, where $d$ is the dimension of the input. We evaluate synthetic perturbations on three benchmark datasets: SVHN \citep{netzer2011reading}, CIFAR-10, CIFAR-100 \citep{cifar}, and a subset of ImageNet \cite{russakovsky2015imagenet}. Following the setup in \citet{huang2021unlearnable}, we use the first 100 classes of the full dataset as the ImageNet subset.  The target model architectures include VGG \citep{simonyan2014very}, ResNet \citep{he2016deep}, and DenseNet \citep{huang2017densely}.  We adopt standard  random cropping and flipping as data augmentation. The  hyperparameters for training are standard and can be found in Appendix~\ref{apdx:exp_details}. We use $\epsilon'=6/255$ for synthetic perturbations. The patch size in Algorithm~\ref{alg:generation} is set as $8$. 

\begin{table}[h]
\caption{Accuracy on clean test data of CIFAR-10. The target model is ResNet-18. The training data are poisoned with different attacks. The closer the accuracy to random guessing, the better the attack efficiency.}

\label{tbl:compare}
\centering

\begin{tabular}{c|c}

Algorithm & Test Accuracy (in \%)  \\ \hline
No Perturbation &  $94.69$   \\
 TensorClog \citep{shen2019tensorclog} &  $48.07$  \\ 
  Alignment \citep{fowl2021preventing} &  $56.65$ \\ 
 DeepConfuse \citep{feng2019learning} &  $28.77$ \\ 
  NTGA \citep{yuan2021neural} &  $33.29$ \\ 
 Error-minimizing \citep{huang2021unlearnable} & $19.93$  \\ 
 Adversarial Examples \citep{fowl2021adversarial} & $6.25$  \\ 
 Synthetic Perturbations  &   $13.54$ \\
\end{tabular}
\end{table} 

We first compare synthetic perturbations with existing poisoning attacks. The comparisons are made on the CIFAR-10 dataset with ResNet-18 as the target model. We use the best-performing setup in their official implementations to generate perturbations. We present the comparison in Table~\ref{tbl:compare}. Synthetic perturbations are as powerful as advanced poisoning attacks despite they are much easier to generate.

\begin{table*}
\renewcommand{\arraystretch}{1.25}
\caption{Accuracy (in \%) on clean test data. The target models are trained on clean data ($\sD_{c}$) and data perturbed by synthetic perturbations ($\sD_{syn}$).}

\label{tbl:different_datasets}
\centering

\begin{threeparttable}
\begin{tabular}{c|cc|cc|cc|cc}

\hline
\multirow{2}{*}{\textbf{\begin{tabular}[c]{@{}c@{}}Target Model\end{tabular}}} &  \multicolumn{2}{c|}{\textbf{SVHN}} & \multicolumn{2}{c|}{\textbf{CIFAR-10}} & \multicolumn{2}{c|}{\textbf{CIFAR-100}}  & \multicolumn{2}{c}{\textbf{ImageNet Subset}} \\ \cline{2-9}
  & \textbf{$\sD_{c}$} &  \textbf{$\sD_{syn}$} &
  \textbf{$\sD_{c}$} &  \textbf{$\sD_{syn}$} &
  \textbf{$\sD_{c}$} &  \textbf{$\sD_{syn}$} & \textbf{$\sD_{c}$} &  \textbf{$\sD_{syn}$}   \\ \hline

VGG-11 & 95.4  & 18.1       &   91.3  & 28.3 & 67.5 &   10.9  &  79.1 &  10.7                    \\ \hline
ResNet-18 & 96.2 & 8.0      & 94.7  & 13.5 & 74.8 & 9.0  & 79.7 &   11.0     \\ \hline
ResNet-50 & 96.4 & 7.8      &   94.8   & 14.9 & 75.2 & 8.4  & 82.4 &    10.8    \\ \hline
DenseNet-121 & 96.7 & 9.7   & 95.0   & 10.6 & 76.5 & 7.6   & 82.9 &     14.7  \\ \hline 
\end{tabular}
\end{threeparttable}

\end{table*}

\begin{figure*}
\centering
  \includegraphics[width=.9\linewidth]{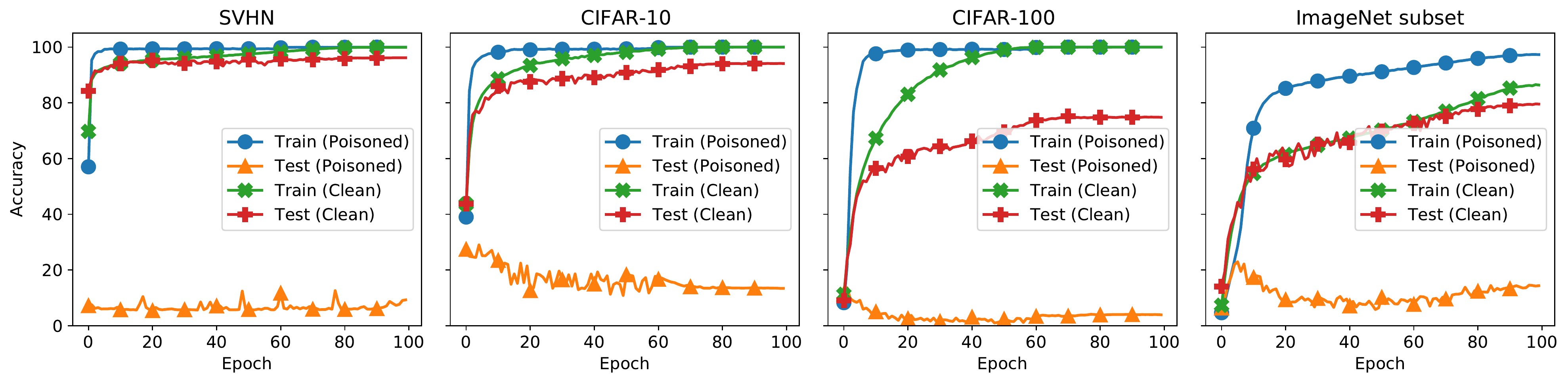}
  \caption{Training curves of ResNet-18 models on perturbed and clean data. The word `poisoned' denotes the model is trained on perturbed data. The test performance is evaluated on clean data. The test accuracy is low throughout training when synthetic perturbations are added. }  
  \label{fig:trn_curve}
\end{figure*}

Then we evaluate synthetic perturbations on different models and datasets. The test accuracy of target models is in Table~\ref{tbl:different_datasets}. We also plot the training curves of target models on both clean and perturbed data in Figure~\ref{fig:trn_curve}. The results in Table~\ref{tbl:different_datasets} and Figure~\ref{fig:trn_curve} further confirm the effectiveness of synthetic perturbations.

We note that generating synthetic perturbations is data irrelevant and only  takes several seconds using a single CPU core.  In contrast, existing attacks often need hours or even days to generate perturbations using GPUs. We compare the computational complexities of  synthetic perturbations and recent attacks in Section~\ref{subsec:cost_comparison}.

In summary, our experiments demonstrate that using linearly-separable perturbations is indeed a sufficient condition for availability poisoning attacks to succeed. Moreover, these results also expose that deep models are very vulnerable to obscured shortcuts. This finding has two meanings to the community. First, it confirms that advanced availability poisoning attacks do succeed by providing shortcuts. Second, it  further exposes the shortcut learning problem, which is a fundamental vulnerability of deep models.

\subsection{Complexity Analysis}
\label{subsec:cost_comparison}

Here we give an analysis of the computational complexity of Algorithm~\ref{alg:generation}. The complexity of generating synthetic perturbations is $\mathcal{O}(nd/p^{2})$, where $n$ is the size of the dataset, $d$ is the dimension of clean data, and $p$ is the patch size. This complexity is mainly from introducing covariance into synthetic data (Line 6 in Algorithm~\ref{alg:generation}). 

The complexity of generating synthetic perturbations is significantly smaller than those of recent attacks. The complexity of running the algorithms in recent attacks is $\mathcal{O}(nTL(dw+w^{2}))$, where $T$ is the number of iterations generating the poisons, $L$ is the network depth, and $w$ is the network width (for simplicity, we assume the network is of equal width). The term $nL(dw+w^{2})$ is the cost of one forward/backward pass. This complexity is strictly worse than that of generating synthetic perturbations.

In recent attacks, the number of iterations generating the poisons is usually large. For example, \citet{huang2021unlearnable} use multiple gradient descent steps to solve the optimization problems in Eq~(\ref{eq:error_min}). On the ImageNet dataset, they first run 100 SGD updates for the outer problem. Then they loop over every target example to optimize the inner problem. They run  20 SGD updates for each example. The above process is repeated until the training accuracy is larger than a pre-defined threshold. Another example is the attack in \citet{fowl2021adversarial}. For each target example, they use 250 Projected Gradient Descent (PGD) steps to generate perturbations.




We now give an empirical comparison. We measure the time costs of generating error-minimizing noises, adversarial examples, and synthetic perturbations. The device is a server with a single Tesla V100 GPU and an Intel Xeon E5-2667 CPU. We note that running Algorithm~\ref{alg:generation} does not require a GPU. The time costs are tested on SVHN, CIFAR-10, and the ImageNet subset. The target model is ResNet18. We use the configurations described in \citet{huang2021unlearnable} and \citet{fowl2021adversarial} to generate error-minimizing noises and adversarial examples.  For synthetic perturbations, we use the same setting as that in Section~\ref{subsec:comparison}. The time costs are reported in Table~\ref{tbl:cost}. Generating synthetic perturbations is significantly cheaper than existing availability attacks.

\begin{table}
\caption{Time costs (in seconds) of generating error-minimizing noises, adversarial examples, and synthetic perturbations.}

\label{tbl:cost}
\centering

\begin{tabular}{c|ccc}
\hline
\hline
Method & SVHN  & CIFAR-10 &   ImageNet     \\ \hline
Error-min. Noises & $\sim$2.7k  & $\sim$3.5k &   $>$28k   \\ \hline
Adv. Examples & $\sim$3.3k  & $\sim$4.1k &   $>$30k   \\ \hline
Algorithm~\ref{alg:generation} & $<$3 & $<$3  & $<$3   \\ \hline \hline
\end{tabular}
\end{table}

\section{Experiments under Different Settings}


Here we test synthetic  perturbations under different settings. We first consider two cases where not all the training data are poisoned. Although the main setting in previous works is to perturb the full training set, in practice we may only need to perturb part of the data. In the first case, we only apply Algorithm~\ref{alg:generation} to some of the classes. In the second case, we perturb partial data that are randomly sampled from all the classes. Finally, we run experiments on a face dataset following the application scenario in \citet{huang2021unlearnable}.


\subsection{Poisoning Some Classes of The Training Data}
\label{subsec:singleclass}



In many practical datasets, some classes are more sensitive than  others, e.g., in medical datasets, the patients that are diagnosed with a certain disease may be more concerned about their data than healthy people. We randomly sample some classes of the CIFAR-10 dataset and apply  Algorithm~\ref{alg:generation} on all the examples of the sampled classes. After training, we report the test accuracy on clean classes and poisoned classes separately. 

  
  


The synthetic perturbations are generated in the same way as that in Section~\ref{subsec:comparison}. For comparison, we also run experiments with error-minimizing noises using those generated in Section~\ref{subsec:comparison}. The experiments are run on ResNet-18. The numbers of poisoned classes are 1, 3, and 5. The poisoned classes are randomly chosen and are the same for two types of noises. We report the results in  Table~\ref{tbl:some_class}.  Algorithm~\ref{alg:generation} is still highly effective when only some of the classes are poisoned.

\begin{table}

\centering
\caption{Test accuracy (in \%) on clean ($\sC$) and poisoned  ($\sP$) classes. Numbers of poisoned classes are 1, 3, and 5.}
\label{tbl:some_class}
\begin{tabular}{c|cc|cc|cc}
\hline
\hline
\multirow{2}{*}{\begin{tabular}[c]{@{}c@{}}Method\end{tabular}} &  \multicolumn{2}{c|}{1}  &  \multicolumn{2}{c|}{3} &  \multicolumn{2}{c}{5}\\ \cline{2-7}
  & $\sC$ & $\sP$ 
  & $\sC$ & $\sP$
  & $\sC$ & $\sP$ \\ \hline
  
 Error-min. Noises \cite{huang2021unlearnable} & 94.6 & 2.4 & 93.9 & 1.1 & 93.8 & 3.1 \\\hline  
  
Synthetic Perturbations & 94.7 & 2.7 & 94.0 & 0.6 & 93.2 & 2.9 \\\hline \hline

\end{tabular}
\end{table}

\subsection{Poisoning Different Percentages of The Training Data}
\label{apdx:different_percentage}

Here we show synthetic perturbations remain effective when only a given percentage of the training data is poisoned. We follow the experimental setup in \citet{huang2021unlearnable,fowl2021adversarial}. Specifically, for each poisoning percentage, we train two models.  One model uses both the clean subset and the poisoned subset as its training data and the other one only uses the clean subset. The difference between the performances of those two models  represents how much information the former model gains from the poisoned data. A small performance gap indicates the former model gains little information from the poisoned data.

We test four different poisoning percentages (from 20\% to 90\%) on the CIFAR-10 dataset. The experiments are run on  ResNet-18 models. We compare the performance of synthetic perturbations with that of adversarial examples and error-minimizing noises \citep{huang2021unlearnable,fowl2021adversarial}. The results are presented in Table~\ref{tbl:different_percentage}. The performance gain of using the poisoned subset is small for all three attacks. This suggests that synthetic perturbations are still effective in this setting.

\begin{table}[t]
\caption{Test accuracy (in \%) with different poisoning percentages $p$.  Training with the poisoned subset does not improve the test accuracy much compared to training with clean data only.}

\label{tbl:different_percentage}
\centering
\renewcommand{\arraystretch}{1.15}
\begin{tabular}{c|cccc}

Method & $p=$90\%  & $p=$80\%   & $p=$50\%   & $p=$20\%    \\ \hline
Clean Data ($1-p$) &   82.6   &  86.5  & 92.4 & 93.9  \\ 
 Error-min. Noises \citep{huang2021unlearnable} & 85.2  & 86.8  & 92.8 &  94.1  \\ 
Adv. Examples \citep{fowl2021adversarial} &  85.3  &  88.2 &  92.2  & 93.7 \\ 
 Synthetic Perturbations & 85.7  & 86.3   & 92.9    & 94.0 \\ 
\end{tabular}
\end{table}


\begin{figure*}[h]
\centering
  \includegraphics[width=0.6\linewidth]{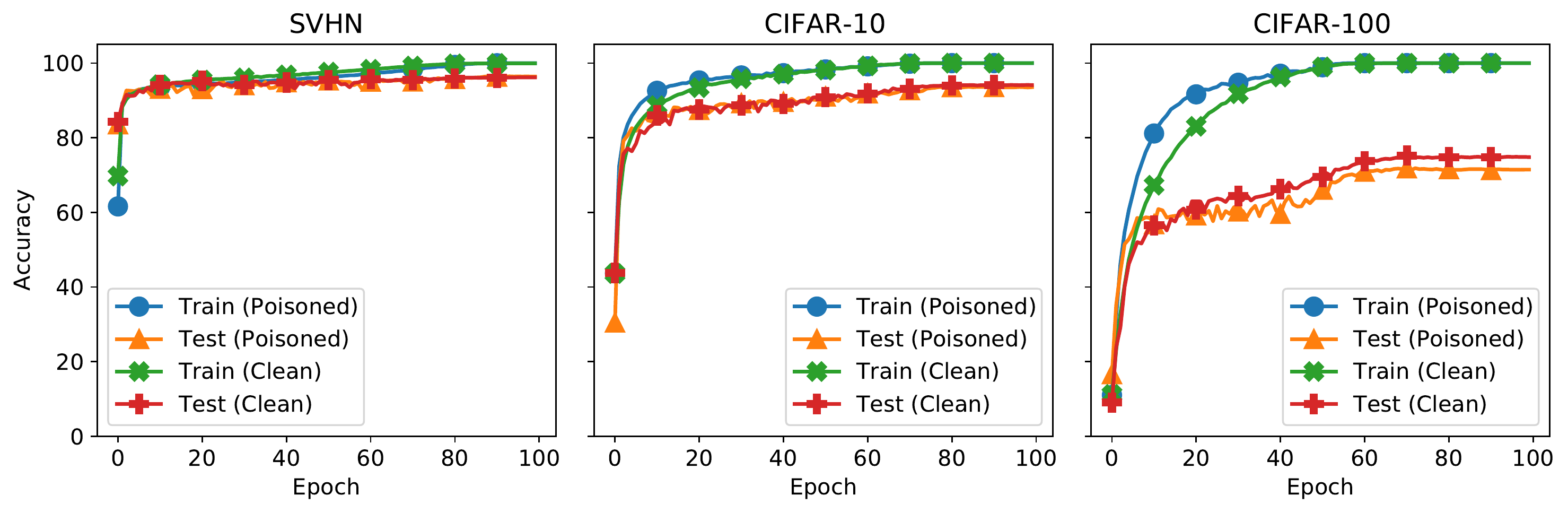}
  \caption{Training curves of ResNet18 models trained on SVHN, CIFAR-10, and CIFAR-100 datasets. The perturbations are \textbf{NOT} processed by the padding opeartion. }
  \label{fig:nopad_trn_curve}
\end{figure*}

\subsection{Experiments on Face Data}
\label{subsec:face}

Here we apply Algorithm~\ref{alg:generation} on face datasets. We follow the application scenario in \citet{huang2021unlearnable} (see Figure 4 in \citet{huang2021unlearnable} for an illustration). The task is to use face images to predict biological identities.   We train an Inception-ResNet-v1 model \citep{szegedy2017inception} on the WebFace Dataset \citep{yi2014learning}. A random subset with 20\%  samples is used for testing and the remaining samples are used for training. The WebFace dataset has 10575 identities and 50 of them are poisoned. We run Algorithm~\ref{alg:generation} with the configuration in Section~\ref{subsec:comparison} ($\epsilon'=6/255$) to process the training images of the poisoned identities.   

\begin{figure}
\centering
\includegraphics[width=0.3\textwidth]{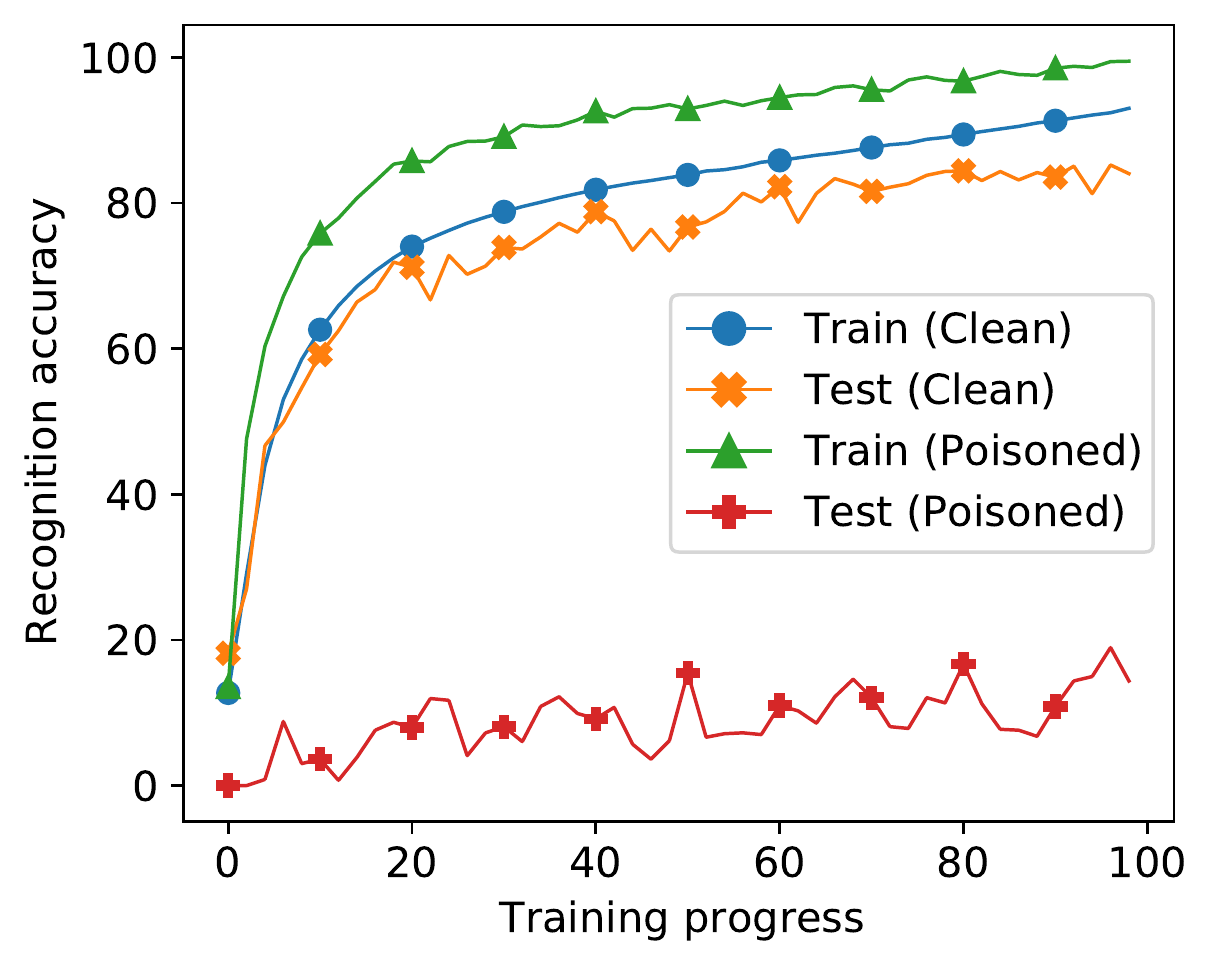}
\caption{Training curves of an Inception-ResNet model on clean/poisoned WebFace.  }
\label{fig:face_training_curve}
\end{figure}

When using Algorithm~\ref{alg:generation}, the test accuracy of the poisoned identities is only 13.6\% which is much lower than the test accuracy of the clean identities (>80\%). The training curves are plotted in Figure~\ref{fig:face_training_curve}.  When using error-minimizing noises, the test accuracy of the poisoned identities reported in \citet{huang2021unlearnable} is $\sim$16\%.  These results confirm that Algorithm~\ref{alg:generation} is also highly effective on face data.  We note that generating error-minimizing noises requires some auxiliary data. For example, \citet{huang2021unlearnable} use 100 identities from the CelebA dataset \citep{liu2015faceattributes} to generate error-minimizing noises for the 50 identities from the WebFace dataset. On the contrary, running Algorithm~\ref{alg:generation} does not require any auxiliary data.

\section{Ablation Study}
\label{sec:abla}

In this section, we run some experiments to better understand our findings and the proposed algorithm. We first test whether the linear separability is stemming from the property of the ReLU activation function. Then we demonstrate the padding operation in Algorithm~\ref{alg:generation} is necessary to make the perturbations robust against data augmentation methods.

\subsection{Linear Separability Is Not Stemming from ReLU}
\label{apdx:tanh}

It is well known to the community that a ReLU-DNN learns a piecewise linear function in input space \citep{arora2016understanding}. The crafting models of advanced availability attacks all use the ReLU activation by default. Therefore, the linear separability may be stemming from the property of ReLU. To verify this, we replace the ReLU layers with Tanh layers in the crafting models of error-minimizing noises \citep{huang2021unlearnable} and adversarial examples \citep{fowl2021adversarial}. We fit the new perturbations with the same simple models as those in Section~\ref{sec:why}. The results are presented in Table~\ref{tbl:linear_acc_tanh}. The new perturbations are still almost linearly separable: linear models achieve more than 90\% training accuracy and two-layer neural networks achieve 100\% training accuracy. This suggests that the linear separability is not associated with the ReLU activation function.

\begin{table}
\caption{Training accuracy (in \%) of simple models on the perturbations of different attacks. The perturbations are generated with \textbf{Tanh-DNNs.}}

\label{tbl:linear_acc_tanh}
\centering

\begin{tabular}{c|cc}

Algorithm & Linear Model & Two-layer NN  \\ \hline
 Error-min. Noises \citep{huang2021unlearnable} & $100.0$ & $100.0$  \\ 
Adv. Examples (Untargeted) \citep{fowl2021adversarial} & $92.7$ & $100.0$ \\ 
 Adv. Examples (Targeted) \citep{fowl2021adversarial} & $100.0$ & $100.0$ \\ 
\end{tabular}
\end{table}

\subsection{The Effect Of The Padding Operation in Algorithm~\ref{alg:generation}}
\label{apdx:no_pad}

Here we explain why we duplicate each dimension of the initial data points into two-dimensional patches in Algorithm~\ref{alg:generation}.  Intuitively, it is more convenient to  directly generate synthetic perturbations that have the same dimension as the original images. We will show  this straightforward approach can not be used as a powerful attack.


We directly use the output of the method in \citet{guyon2003design} as the straightforward approach, i.e., the dimension of synthetic data is the same as the dimension of flattened images and we simply reshape the synthetic data into the image format. Other configurations are the same as those in Section~\ref{sec:syntehtic_noise}. The models are trained with standard augmentation methods including random crop and flipping.  The training curves of the target models are plotted in Figure~\ref{fig:nopad_trn_curve}. The test accuracy is still high when the data is poisoned, which does not meet the requirement of availability attacks. 

The fact that the padding operation makes the perturbations remain effective may be because it introduces local correlation into the perturbations, which is an inherent property of natural images. In Appendix~\ref{apdx:more_aug}, we show synthetic perturbations are still highly effective when more powerful data augmentation methods  are applied.

\section{Conclusion}

This work gives an explanation of the working principle of availability poisoning attacks. We show advanced attacks coincidentally  generate linearly-separable perturbations. We further synthesize linearly separable  perturbations to demonstrate that using linearly separable perturbations is sufficient for an availability attack to succeed. The proposed algorithm is an order of magnitude faster than existing attacks. Our findings also suggest deep models are more prone to shortcuts than previously believed. They will find and heavily rely on shortcuts even when the shortcuts are scaled down to an imperceptible magnitude.

\section*{Acknowledgement}

Da Yu and Jian Yin are with Guangdong Key Laboratory of Big Data Analysis and Processing, School of Computer Science and Engineering, Sun Yat-sen University. Jian Yin is supported by the National Natural Science Foundation of China (U1811264, U1811262, U1811261, U1911203, U2001211), Guangdong Basic and Applied Basic Research Foundation (2019B1515130001), Key-Area Research and Development Program of Guangdong Province \\(2018B010107005, 2020B0101100001).




\bibliography{acmart}
\bibliographystyle{ACM-Reference-Format}

\begin{appendices}

\begin{appendix}

\section{Additional t-SNE Plots}
\label{apdx:more_tsne}

\begin{figure}[h]
\centering
  \includegraphics[width=0.85\linewidth]{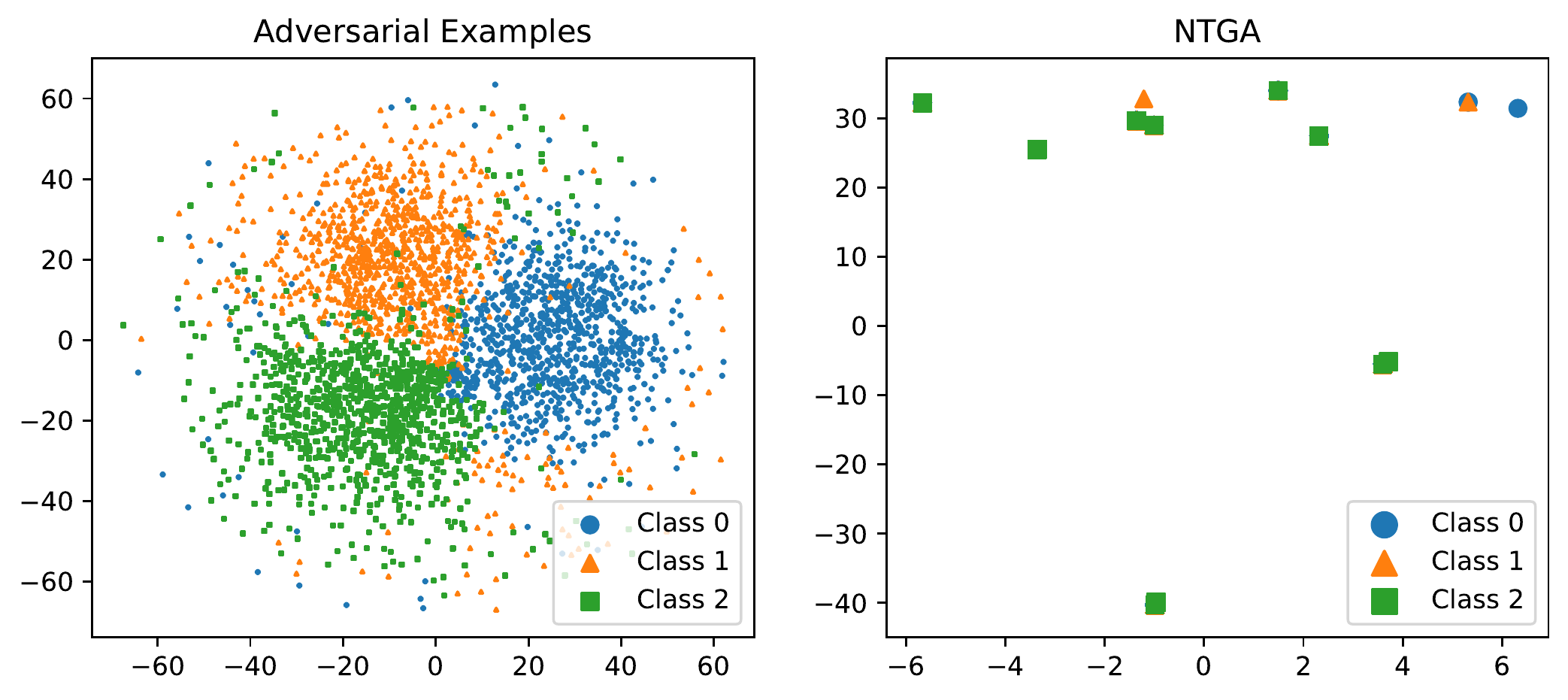}
  \caption{T-SNEs of targeted adversarial examples \citep{fowl2021adversarial} and NTGA \citep{yuan2021neural}. Perturbations from the same class are well clustered. Notably, many embeddings of NTGA are overlapped. }
  \label{fig:add_tsne}
\end{figure}

Here we plot the t-SNEs of two other attacks in Table~\ref{tbl:linear_acc}, i.e., adversarial examples \citep{fowl2021adversarial} and NTGA \citep{yuan2021neural}. We use their official implementations to generate the perturbations (see Appendix~\ref{apdx:exp_details} for details).   The t-SNEs are plotted in Figure~\ref{fig:add_tsne}. The perturbations for the same class are well clustered. This observation is similar to  that from Figure~\ref{fig:tsne}.

\section{Implementation Details}
\label{apdx:exp_details}

\textbf{Implementation details of the experiments in Section~\ref{sec:why}.} We generate perturbations of baseline algorithms using their official implementations:  DeepConfuse\footnote{\url{https://github.com/kingfengji/DeepConfuse}}, NTGA\footnote{\url{https://github.com/lionelmessi6410/ntga}},  error-minimizing noise\footnote{\url{https://github.com/HanxunH/Unlearnable-Examples}}, and adversarial examples\footnote{\url{https://github.com/lhfowl/adversarial_poisons}}. The configuration is set to be the one that achieves the best attack performance on CIFAR-10. Specifically, DeepConfuse uses an 8-layer U-Net \citep{ronneberger2015u} as the crafting model. NTGA uses a 3-layer convolutional network. Error-minimizing noises and adversarial examples use standard ResNet-18 models. 

The experimental setup for training the simple models is as follows. We train the simple models  with standard cross-entropy loss. Before training, all perturbations are flattened into 1-dimensional vectors and normalized to unit norm. The two-layer neural networks have a width of $30$. All models are trained with the L-BFGS optimizer \citep{liu1989limited} for 50 steps.



\textbf{Implementation details of the experiments in Section~\ref{sec:syntehtic_noise}.} We use the Stochastic Gradient Descent (SGD) optimizer with a momentum coefficient 0.9 for all experiments. For all datasets, we use a batchsize of 128. The learning rates of all models are set to follow the choices in the original papers \citep{simonyan2014very,he2016deep,huang2017densely}. The learning rate for ResNet and DenseNet models is 0.1. The learning rate for VGG models is 0.01.  All models are trained for 100 epochs. The learning rate is divided by 10 at epoch 50 and 75.

\section{Training with More Data Augmentation Methods}
\label{apdx:more_aug}

Here we demonstrate that synthetic noises can not be filtered out by state-of-the-art data augmentation methods. We test  four advanced data augmentation methods including Cutout \citep{devries2017improved}, Mixup \citep{zhang2017mixup}, CutMix \citep{yun2019cutmix}, and Fast Autoaugment (FA) \citep{lim2019fast}. Experimental results suggest that  synthetic perturbations are still highly effective when those augmentation methods are applied.


\begin{table}
\caption{Test accuracy (in \%) of ResNet18 models on the CIFAR-10 dataset.}
\renewcommand{\arraystretch}{1.25}  
\label{tbl:acc_aug}
\centering
\small
\begin{tabular}{c|c|c|c|c}
\hline
\hline
  & Cutout &   Mixup   & CutMix &  FA      \\ \hline  
  
Error-min. Noises  & 18.9 &  57.4  & 32.3 &  41.6          \\ \hline
Synthetic Perturbations  & \textbf{10.6}  & \textbf{39.5}    & \textbf{17.7} & \textbf{24.4}   \\ \hline \hline

\end{tabular}
\end{table}

We train ResNet18 models on the CIFAR-10 dataset.  For all augmentation methods, we use the default configurations for CIFAR-10 from the original papers to set their parameters. Other experimental settings such as the noise strength and training recipe are the same as those in Section~\ref{subsec:comparison}. The results are presented in Table~\ref{tbl:acc_aug}. In Table~\ref{tbl:acc_aug}, we also include the test accuracy of using error-minimizing noises for a comparison. The results suggest that   synthetic 
perturbations are more effective when advanced augmentation methods are applied. For example, when Fast Autoaugment (FA) is applied, the test accuracy of using synthetic perturbations is 24.4\% while the test accuracy of using error-minimizing noises is 41.6\%.


\end{appendix}
\end{appendices}

\end{document}